\documentclass[sigconf]{acmart}
\usepackage{float}
\usepackage{tabularray,varwidth,enumitem}
\usepackage{array}
\usepackage{amsmath}
\DeclareMathOperator*{\argmax}{argmax}

\usepackage{graphics}
\usepackage{wrapfig}
\usepackage{multirow, makecell}
\usepackage{multicol}
\usepackage{graphicx}
\usepackage{caption}
\usepackage{lipsum} % for dummy text
\usepackage{subcaption}
\AtBeginDocument{%
  }

\setcopyright{acmlicensed}

\copyrightyear{2024}
\acmYear{2024}
\setcopyright{acmlicensed}\acmConference[MMASIA '24]{ACM Multimedia
Asia}{December 3--6, 2024}{Auckland, New Zealand}
\acmBooktitle{ACM Multimedia Asia (MMASIA '24), December 3--6, 2024,
Auckland, New Zealand}
\acmDOI{10.1145/3696409.3700211}
\acmISBN{979-8-4007-1273-9/24/12}

\begin{document}

\title{Action Selection Learning for \\ Multi-label Multi-view Action Recognition}
\renewcommand{\shorttitle}{Action Selection Learning for Multi-label Multi-view Action Recognition}

% \author{No Author Given}

\author{Trung Thanh Nguyen}
\email{nguyent@cs.is.i.nagoya-u.ac.jp}
\orcid{0000-0001-8976-2922}
\affiliation{%
  \institution{Nagoya University, Nagoya, Japan}
  % \city{Nagoya}
  % \state{Aichi}
  % \country{Japan}
  \country{}
}
\affiliation{%
  \institution{RIKEN, Kyoto, Japan}
  % \city{Seika}
  % \state{Kyoto}
  % \country{Japan}
  \country{}
}

\author{Yasutomo Kawanishi}
\email{yasutomo.kawanishi@riken.jp}
\orcid{0000-0002-3799-4550}
\affiliation{%
  \institution{RIKEN, Kyoto, Japan}
  % \city{Seika}
  % \state{Kyoto}
  % \country{Japan}
  \country{}
}
\affiliation{%
  \institution{Nagoya University, Nagoya, Japan}
  % \city{Nagoya}
  % \state{Aichi}
  % \country{Japan}
  \country{}
}

\author{Takahiro Komamizu}
\email{taka-coma@acm.org}
\orcid{0000-0002-3041-4330}
% \affiliation{%
%   \institution{Mathematical and Data Science Center, Nagoya University}
%   \city{Nagoya}
%   \state{Aichi}
%   \country{Japan}
% }
\affiliation{%
  \institution{Nagoya University, Nagoya, Japan}
  % \city{Nagoya}
  % \state{Aichi}
  % \country{Japan}
  \country{}
}

\author{Ichiro Ide}
\email{ide@i.nagoya-u.ac.jp}
\orcid{0000-0003-3942-9296}
\affiliation{%
  \institution{Nagoya University, Nagoya, Japan}
  % \city{Nagoya}
  % \state{Aichi}
  % \country{Japan}
  \country{}
}
% \affiliation{%
%   \institution{Mathematical and Data Science Center, Nagoya University}
%   \city{Nagoya}
%   \state{Aichi}
%   \country{Japan}
% }

\renewcommand{\shortauthors}{Trung Thanh Nguyen et al.}

\begin{abstract}
Multi-label multi-view action recognition aims to recognize multiple concurrent or sequential actions from untrimmed videos captured by multiple cameras. 
Existing work has focused on multi-view action recognition in a narrow area with strong labels available, where the onset and offset of each action are labeled at the frame-level.
This study focuses on real-world scenarios where cameras are distributed to capture a wide-range area with only weak labels available at the video-level.
We propose a method named \textbf{Multi}-view \textbf{A}ction \textbf{S}election \textbf{L}earning (\textbf{MultiASL}), which leverages action selection learning to enhance view fusion by selecting the most useful information from different viewpoints. 
The proposed method includes a Multi-view Spatial-Temporal Transformer video encoder to extract spatial and temporal features from multi-viewpoint videos. 
Action Selection Learning is employed at the frame-level, using pseudo ground-truth obtained from weak labels at the video-level, to identify the most relevant frames for action recognition.
Experiments in a real-world office environment using the MM-Office dataset demonstrate the superior performance of the proposed method compared to existing methods.
\end{abstract}



% \received{20 February 2024}
% \received[revised]{12 March 2024}
% \received[accepted]{5 June 2024}

\maketitle

\section{Introduction}
\label{sec:introduction}
\begin{figure}[t]
    \centering
    \includegraphics[width=0.475\textwidth]{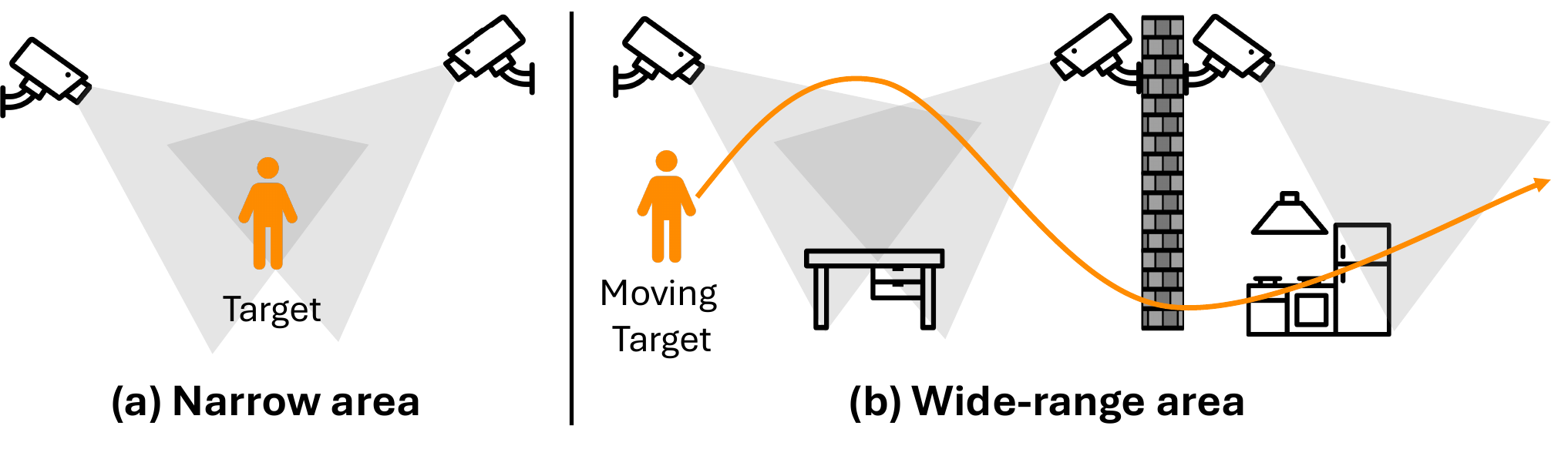}
    % \vspace{-15pt}
    \caption{Configuration of multi-camera settings. 
    (a) Multiple cameras arranged to surround a target in a narrow area. (b)~Multiple distributed cameras covering a wide-range area, which is the environment targeted in this study.}
    \label{fig:multi_view_settings}
    \vspace{-10pt}
\end{figure}

Action recognition is an active research area in computer vision, garnering significant interest due to its wide range of applications, including surveillance~\cite{khan2024human}, robotics~\cite{voronin2021action}, and video content analysis~\cite{pareek2021survey}. 
With the rapid advancements in multi-camera systems~\cite{olagoke2020literature}, there is an increasing need to capture and analyze actions from multiple viewpoints to obtain a comprehensive understanding. 

Conventional single-view action recognition methods~\cite{gupta2021quo, sun2022human, kong2022human} are inherently limited by the perspective from which the action is observed, often resulting in incomplete understanding and potential misclassification of actions. 
In contrast, multi-view action recognition approaches integrate information from different viewpoints, providing a more holistic understanding of actions through complementary information. 
However, as shown in Figure~\ref{fig:multi_view_settings}a, current research~\cite{wang2021continuous, bai2020collaborative, shah2023multi, 10.1007/978-3-030-58583-9_26} primarily focuses on multi-camera setups surrounding a target in a narrow area. 
This setup is constrained for real-world scenarios where a target can move across a wide-range area.
Figure~\ref{fig:multi_view_settings}b illustrates cameras distributed to capture a wide-range area.
The success of multi-view action recognition lies in the effective joint representation of complementary information from different viewpoints. 
The challenge in a wide-range area is fusing information from various cameras, especially when some cameras may capture irrelevant information if the target is outside their view.

Typically, deep learning models for action recognition achieve the best results with strong labels, where the onset and offset of each action are annotated at the frame-level. 
However, such detailed annotation is costly for multiple cameras, and in many cases, only weak labels in the form of multi-label tags at the video-level are available~\cite{yasuda2022multi, yasuda2024guided}. 
While some works~\cite{ijcai2018p375, Zhao_9939043} have explored multi-label multi-view action recognition, they have primarily focused on multi-view setups in narrow areas, and the challenges of using weak labels still need to be adequately addressed. 
Unlike previous works, this study addresses the challenges of action recognition in wide-range areas by focusing on multi-view fusion and multi-label recognition based on weak labels. 
We propose the MultiASL method, which employs action selection learning for  multi-view action recognition to identify the most relevant actions using only video-level labels.
This study makes the following contributions:

\begin{itemize}
    \item We propose the Multi-view Spatial-Temporal Transformer video encoder to extract spatial and temporal features from videos, combined with Action Selection Learning (ASL) to identify useful frames for integrating information from multiple views using only video-level labels.
	\item We explore various multi-view fusion strategies (e.g., max pooling, mean pooling, sum, and concatenation) to determine the most effective fusion method. 
    The findings show that max pooling consistently yields the best results.
    \item Experiments on real-world office environments using the MM-Office dataset~\cite{yasuda2022multi} demonstrate the superior performance of the proposed MultiASL method compared to existing methods.
\end{itemize}

\section{Related Work}
\label{sec:related_work}

\noindent \textbf{Action Recognition.}
Advancements in action recognition have leveraged various deep learning techniques to improve the accuracy and robustness of identifying human actions in videos. 
Compared to image classification~\cite{elngar2021image}, action recognition depends on spatiotemporal features extracted from consecutive frames. 
Early attempts at video understanding used a combination of 2D Convolutional Neural Network (CNN)~\cite{wang2022temporal, shen20212d, kim2020action} or 3D CNN~\cite{chen2021deep, alfaifi2020human, wu2021spatiotemporal} to capture spatial and temporal information. 
Recently, with the success of Transformers~\cite{vaswani2017attention} leading to the development of Vision Transformers (ViT)~\cite{dosovitskiy2020image}, researchers have proposed Transformer-based and ViT-based models for action recognition~\cite{hong2023fluxformer, doshi2023semantic, bertasius2021space, arnab2021vivit} that effectively capture long-range spatiotemporal relationships, surpassing their convolutional counterparts.
% However, all the methods discussed above primarily address single-view videos and do not account for multi-view action recognition.

\noindent \textbf{Multi-view Action Recognition.}
The development of datasets with multi-view and multiple modalities (i.e., RGB, depth, and skeleton)~\cite{shahroudy2016ntu, liu2019ntu} has enabled recent progress in multi-view action recognition. 
Most works based on Skeleton methods~\cite{shi2020decoupled, 9710612, zhang2019view} have received significant attention due to the availability of accurate 3D skeleton ground-truth. 
However, estimating an accurate 3D skeleton from videos without access to depth information is challenging. 
Besides that, to avoid dependence on the skeleton modality, some studies focus on RGB-based multi-view methods, proposing supervised contrastive learning~\cite{shah2023multi} or an unsupervised representation learning framework~\cite{10.1007/978-3-030-58583-9_26} to learn a feature embedding robust to changes in viewpoint. 
However, current works typically utilize multiple views in a narrow area.
% Despite these advances, current works typically utilize multiple views arranged to surround humans. In this study, we focus on the task of multi-view action recognition where the cameras are distributed to cover wide-ranging areas.

\noindent \textbf{Multi-label Action Recognition.}
Recent research involves recognizing multiple actions performed simultaneously or sequentially in videos~\cite{yeung2018every, ray2018scenes}. Most studies focus on single-action recognition, employing a sigmoid activation function for the output to provide multi-label predictions~\cite{feichtenhofer2019slowfast, zhang2021multi}. Other works introduce an actor-agnostic approach with a multi-modal query network to enhance multi-label action recognition~\cite{mondal2023actor}. Semi-supervised learning methods address class imbalance in multi-label classification~\cite{zhang2022semi}, while weakly-supervised learning aids in recognizing multiple actions with weak labels~\cite{wang2020multi}. These advancements emphasize the importance of robust methods for handling multiple concurrent actions in diverse video datasets.

\noindent \textbf{Multi-label Multi-view Action Recognition.}
While existing works have significantly contributed to multi-label and multi-view learning, the challenge of using weak labels has yet to be adequately addressed in~\cite{ijcai2018p375, Zhao_9939043}. In these cases, videos have only video-level annotations without frame-level annotations.
Moreover, in the context of distributed cameras covering a wide-range area, irrelevant views can significantly impact recognition accuracy when fusing information from multiple views. 
To address this challenge, we propose action selection learning at the frame-level using only video-level label for multi-view action recognition to reduce irrelevant information and enhance accuracy.

\section{Methodology}
\label{sec:methodology}

\begin{figure*}[t]
    \centering
    % First subfigure
    \begin{subfigure}[b]{0.58\textwidth}
        \centering
        \includegraphics[width=\textwidth]{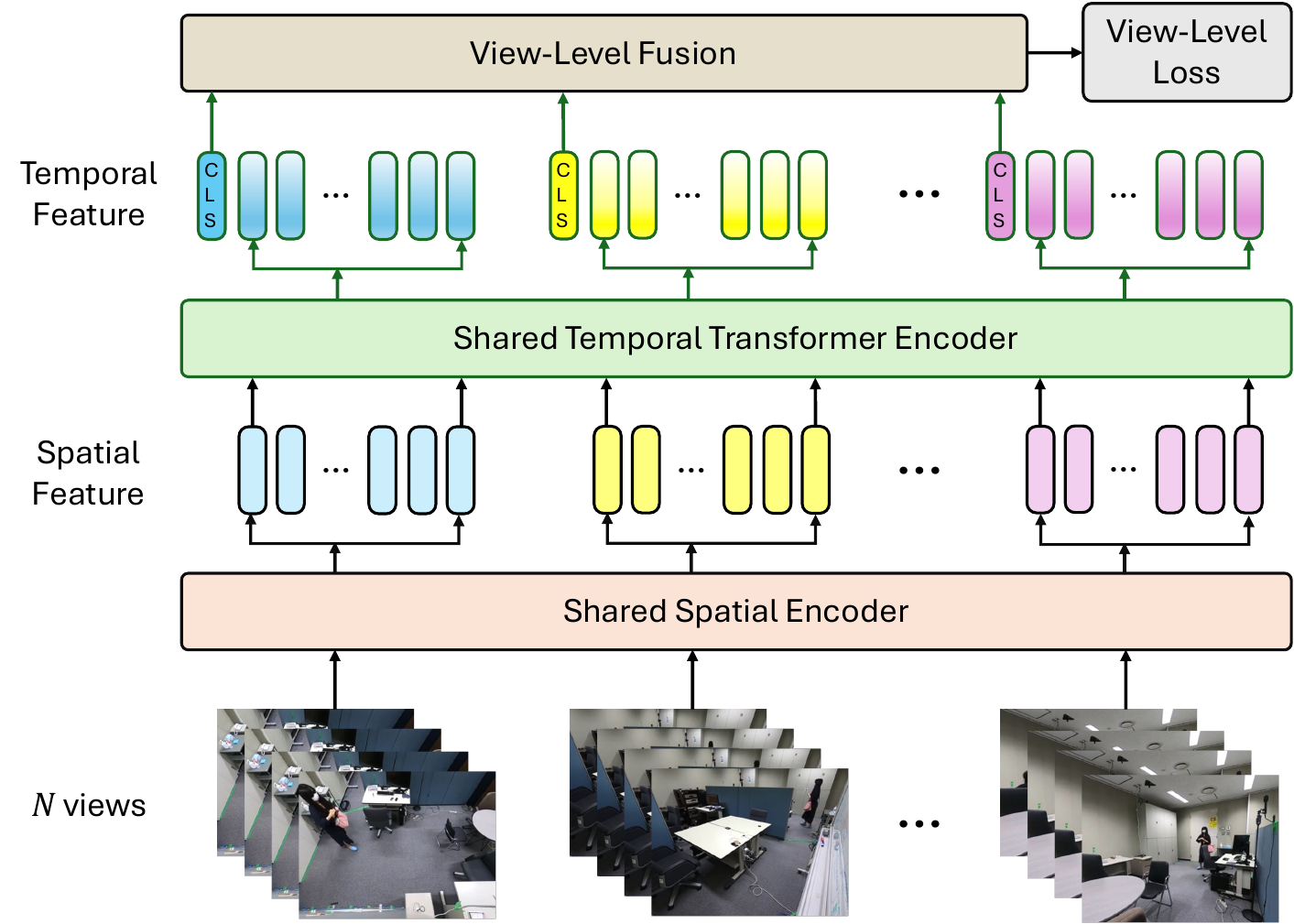}
        \caption{Multi-view Spatial-Temporal Transformer Video Encoder.}
        \label{fig:overview}
    \end{subfigure}
    \hfill 
    \begin{subfigure}[b]{0.375\textwidth}
        \centering
        \includegraphics[width=\textwidth]{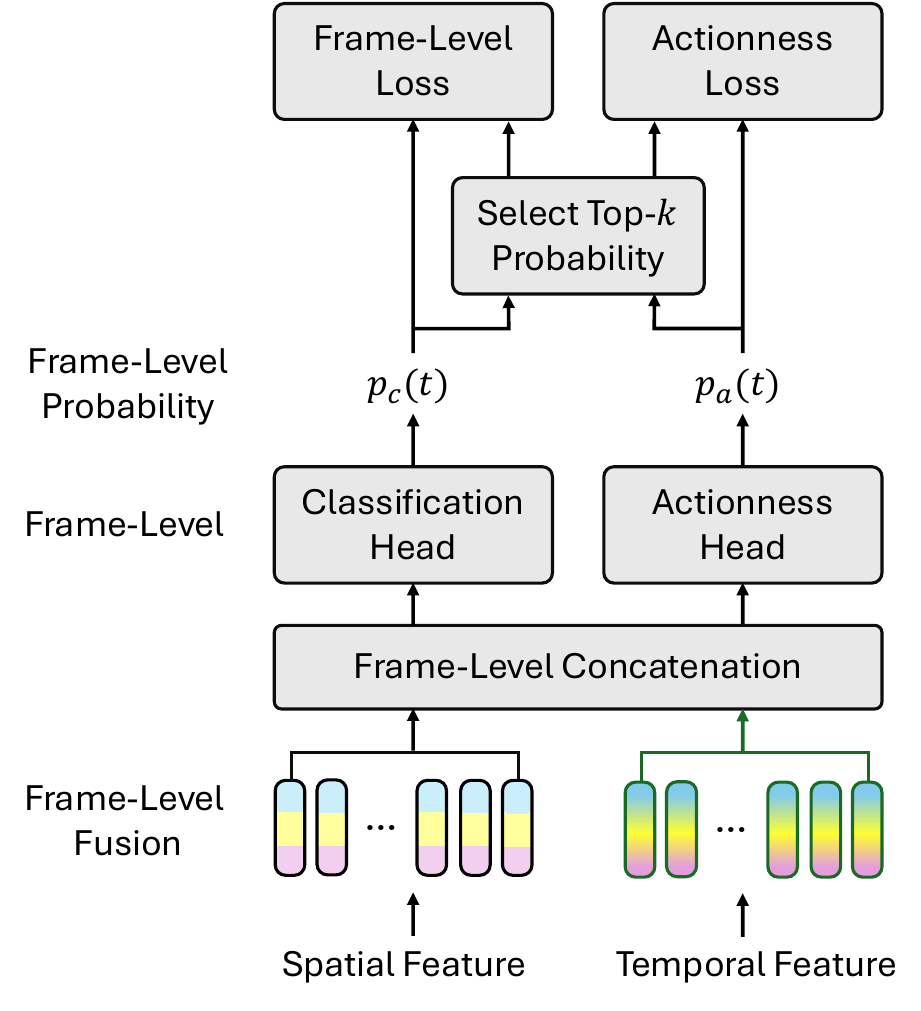}
        \caption{Action Selection Learning.}
        \label{fig:ASL}
    \end{subfigure}
    \caption{Overview of the proposed MultiASL model. 
    It takes videos from \(N\) different views as input and predicts multi-label actions. (a)  Each video is processed by a Shared Spatial Encoder to extract spatial features, which are then fed into a Shared Temporal Transformer Encoder to capture temporal dependencies and generate temporal features. Finally, view-level features are aggregated for action recognition. (b) Frame-level spatial and temporal features are fused to select actions based on video-level labels.}
    \label{fig:multiASL_method}
    \vspace{-5pt}
\end{figure*}

Figure~\ref{fig:multiASL_method} shows the two main components of the proposed MultiASL method, which is trained in an end-to-end setting: Video Encoder (Figure~\ref{fig:overview}) and ASL (Figure~\ref{fig:ASL}).
The model processes a set of videos $\mathcal{V} = \{V_1, V_2, \dots, V_N\}$ from \(N\) views, represented as \(V_i \in \mathbb{R}^{T \times D \times H \times W}\), where \(T\) is the number of frames, \(D\) is the number of channels, and \(H\) [pixels] and \(W\) [pixels] are the height and width of each frame, respectively. 
The Video Encoder extracts spatial and temporal features to predict actions at the video-level, while ASL uses the frame-level spatial and temporal features to select actions based on video-level labels.
This approach addresses the challenge of irrelevant frames in multi-view feature fusion, outputting multi-label action predictions \(A \in \{0, 1\}^{C}\), where \(C\) is the number of action classes.

\subsection{Video Encoder}
\label{subsec:spatial_temporal}
Figure~\ref{fig:overview} shows an overview of the Video Encoder used to extract spatial and temporal features from videos taken from \(N\) views. 
In the encoder, the Shared Spatial Encoder aims to extract detailed spatial features from each frame of the videos, capturing important visual information. 
This is followed by the Shared Temporal Transformer Encoder, which captures temporal dependencies to understand the sequence and progression of actions over time. 
Finally, the aggregated features from multiple views are used for video-level action prediction. 
Additionally, the frame-level spatial and temporal features are used for ASL.

\subsubsection{Shared Spatial Encoder.}
Each video \(V_i\) from the \(i\)-th view (\(i \in \{1, 2, \ldots, N\}\)) is processed by a spatial feature extractor, such as a CNN (e.g., ResNet~\cite{he2016deep}) or ViT~\cite{dosovitskiy2020image}. 
This extractor processes each of the \(T\) frames of each video, capturing spatial features from each frame. The output of the spatial feature extractor for each view \(i\) is a set of spatial features, denoted as \(F_i^{\mathsf{S}} \in \mathbb{R}^{T \times D_\mathsf{S}}\), where \(D_\mathsf{S}\) is the dimensionality of the extracted features.

\subsubsection{Shared Temporal Transformer Encoder.}
The spatial features \(F_i^{\mathsf{S}} \in \mathbb{R}^{T \times D_\mathsf{S}}\) for each view \(i \in \{1, 2, \ldots, N\}\) are then fed as input to a Temporal Transformer Encoder. 
This encoder is designed to capture temporal dependencies across the frames within each video, enabling the model to understand the sequence and progress of actions over time. 
The Temporal Transformer Encoder leverages self-attention mechanisms~\cite{vaswani2017attention} to weigh the importance of each frame relative to the others, effectively summarizing the temporal dynamics. 
The output of this process is a set of temporal features for each view, denoted as \(F_i^{\mathsf{T}} \in \mathbb{R}^{T \times D_{\mathsf{T}}}\), where \(D_{\mathsf{T}}\) represents the dimensionality of these temporal features. 
Additionally, a classification token (CLS) is used within the transformer architecture to aggregate the information from all frames. 
The CLS token \(F_i^{\text{CLS}} \in \mathbb{R}^{D_{\mathsf{T}}}\) is extracted at the output and is used to represent the entire video for video-level classification.

\begin{figure}[t]
    \centering
    \includegraphics[width=0.41\textwidth]{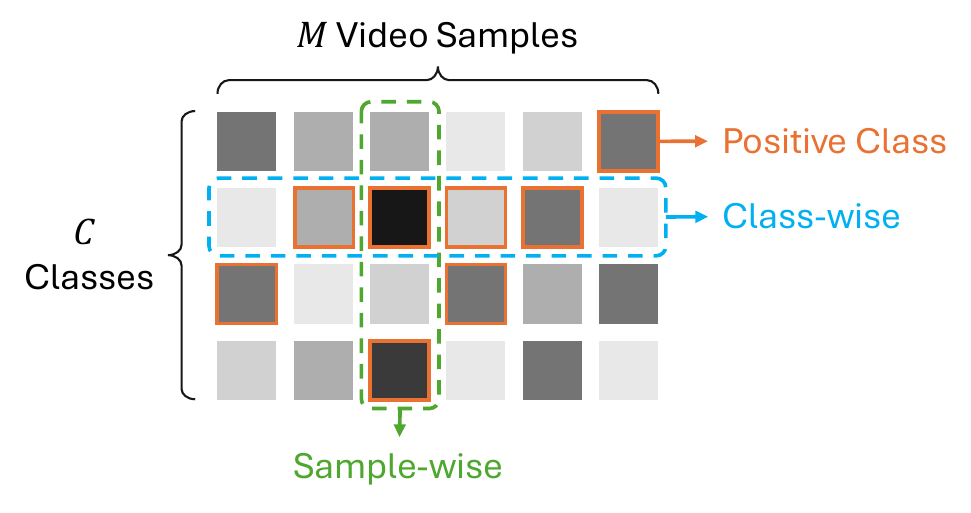}
    \caption{Logit matrix $\mathbf{X}$ for \(M\) video samples and \(C\) classes for Two-way multi-label loss~\cite{kobayashi2023two}.}
    \label{fig:logit_matrix}
    \vspace{-10pt}
\end{figure}

\subsubsection{View-Level Classifier}
We aggregate the video feature representations \(F_i^{\text{CLS}} \in \mathbb{R}^{D_{\mathsf{T}}}\) from each view \(i \in \{1, 2, \ldots, N\}\) using fusion operation (e.g., max pooling, mean pooling, and sum). 
The view-level fused feature \(F_{\text{View-fused}}\) is then passed through a linear layer for view-level classification.
Typically, Binary Cross Entropy (BCE) loss is commonly used in multi-label classification and discriminates samples for each class using a sigmoid function, which can be regarded as \textit{class-wise} classification. 
However, BCE loss faces issues due to class imbalance, leading to suboptimal performance. Inspired by~\cite{kobayashi2023two}, we employ Two-way multi-label loss, which effectively exploits the discriminative characteristics through relative comparisons among classes and video samples.
Figure~\ref{fig:logit_matrix} visualizes the logit matrix $\mathbf{X}$ for \(M\) video samples and \(C\) classes. The two-way loss for multi-label classification incorporates both the \textit{sample-wise} and \textit{class-wise} losses to enhance discrimination in both dimensions. 
The overall View-level loss is defined as:
\begin{equation}
\mathcal{L}_\text{View-level} = \mathcal{L}_{\text{Sample-wise}} + \alpha \mathcal{L}_{\text{Class-wise}},
\label{eq:loss_view_level}
\end{equation}
where \(\alpha\) is a balancing parameter.

\noindent \textbf{Sample-wise loss} focuses on discriminating the positive and negative classes for each video sample as:
\small
\begin{equation}
\nonumber
\mathcal{L}_{\text{Sample-wise}} = \frac{1}{M} \sum_{m=1}^{M} \text{softplus} \left( \log \sum_{n \in \mathcal{N}_m} e^{x_{S_n}} + \gamma \log \sum_{p \in \mathcal{P}_m} e^{-\frac{x_{S_p}}{\gamma}} \right),
\end{equation}
\normalsize
where \(\mathcal{P}_m\) and \(\mathcal{N}_m\) denote the sets of positive and negative labels for the \(m\)-th video sample, respectively. 
Here, \(x_{S_n}\) and \(x_{S_p}\) represent the logits for the positive and negative classes, respectively. 
\(\gamma\) is a temperature parameter.

\noindent \textbf{Class-wise loss} focuses on discriminating the samples within each class as:
\small
\begin{equation}
\nonumber
\mathcal{L}_{\text{Class-wise}} = \frac{1}{C} \sum_{c=1}^{C} \text{softplus} \left( \log \sum_{n \in \mathcal{N}_c} e^{x_{C_n}} + \gamma \log \sum_{p \in \mathcal{P}_c} e^{-\frac{x_{C_p}}{\gamma}} \right),
\end{equation}
\normalsize
where \(\mathcal{P}_c\) and \(\mathcal{N}_c\) denote the sets of samples with positive and negative labels for the \(c\)-th class, respectively. 
Here, \(x_{C_n}\) and \(x_{C_p}\) represent the logits for the positive and negative samples within the class \(c\), respectively. 
\(\gamma\) is a temperature parameter.

\subsection{Action Selection Learning}
\label{subsec:ASL}
Figure~\ref{fig:ASL} shows an overview of ASL. 
The input for ASL consists of the spatial feature \(F_i^{\mathsf{S}} \in \mathbb{R}^{T \times D_{\mathsf{S}}}\) and the temporal feature \(F_i^{\mathsf{T}} \in \mathbb{R}^{T \times D_{\mathsf{T}}}\) from each view \(i \in \{1, 2, \ldots, N\}\). 
Each frame in the spatial and temporal feature is aggregated at the frame level using a fusion operation (e.g., max pooling, mean pooling, and sum). 
The resulting frame-level fused features are \(F_{\text{Frame-fused}}^{\mathsf{S}} \in \mathbb{R}^{D_{\mathsf{S}}}\) and \(F_{\text{Frame-fused}}^{\mathsf{T}} \in \mathbb{R}^{D_{\mathsf{T}}}\). 
These fused spatial and temporal features are then concatenated to obtain the combined feature \(F_{\text{Frame-fused}}^{\mathsf{ST}} \in \mathbb{R}^{D_{\mathsf{ST}}}\), where \(D_{\mathsf{ST}} = D_{\mathsf{S}} + D_{\mathsf{T}}\).

ASL aims to learn to select actions at the frame-level, for which we design two classifiers: Frame-level classifier and Actionness classifier. 
The features \(F_{\text{Frame-fused}}^{\mathsf{ST}}\) are fed into the classification head and actionness head to obtain the frame-level probability of class \(p_c{(t)}\) and actionness \(p_a{(t)}\) for \(t \in \{1, \ldots, T\}\).
The Frame-level classifier aims to predict the video class based on an aggregate of the top-\(k\) frame probabilities using the Multiple Instance Learning (MIL)~\cite{carbonneau2018multiple} approach, with the top-\(k\) probability being the sum of \(p_c{(t)}\) and \(p_a{(t)}\). 
On the other hand, the Actionness classifier aims to predict a binary logit for each frame, indicating whether it is an action or non-action based on the video-level class. 
Since we only have video-level labels, we select the top-\(k\) probabilities of \(p_c{(t)}\) and \(p_a{(t)}\) to serve as the ground-truth for actionness loss.

\subsubsection{Frame-Level Classifier}
For class probability \(p_c{(t)}\) over \(t = \{1, \ldots, T\}\) frames, the MIL approach is commonly employed to train the classifier when only video-level labels are available.
This approach selects the top-\(k\) probabilities for each class to aggregate the highest frame-level probabilities and make video-level predictions.
We denote the set of the top-\(k\) probabilities for each class as:
\small
\begin{equation}
\mathcal{T}^c = \argmax_{\substack{\mathcal{T} \subseteq \{1, \ldots, T\} \\ \: \text{s.t.} \: |\mathcal{T}| = k}}  \sum_{t \in \mathcal{T}} \left(p_{c}(t) + p_a{(t)} \right),
\label{eq:top_k_selection}
\end{equation}
\normalsize
where \(\mathcal{T} \subseteq \{1, \ldots, T\}\) represents a subset of frames from the total \(T\) frames, \(|\mathcal{T}| = k\) means that the subset \(\mathcal{T}\) contains \(k\) frames, \(p_c{(t)}\) is the class probability, and \(p_a{(t)}\) is the actionness probability.
Then the video-level class probability can be selected by aggregation (e.g., mean pooling) using \(\mathcal{T}^c\) as
$p_c = \textsc{Aggregation} \left( \mathcal{T}^c \right)$.
For instance, when using mean pooling, it can be expressed as:
\small
\begin{equation}
p_c = \frac{1}{k} \sum_{t \in \mathcal{T}^c} p_c(t).
\end{equation}
\normalsize
Finally, similar to the View-level loss in Eq.~\ref{eq:loss_view_level}, we employ the Two-way multi-label loss to obtain the Frame-level loss \(\mathcal{L}_{\text{Frame-level}}\).

\begin{figure}[t]
    \centering
    \includegraphics[width=0.4\textwidth]{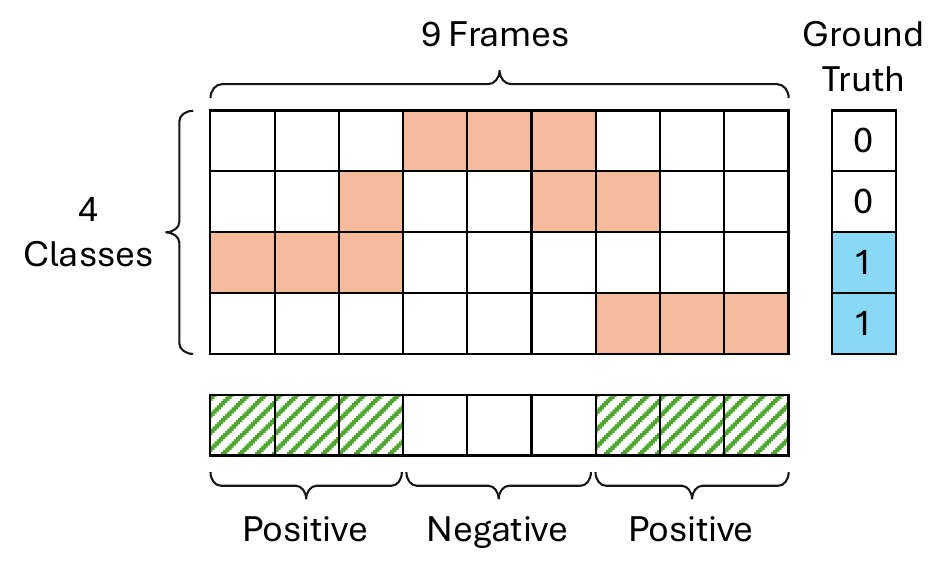}
    \caption{Example of generating pseudo ground-truth for actionness loss. 
    The video consists of 9 frames and 4 video-level classes, with the ground-truth classes being 3 and 4. We select the top-3 predictions for each class (in orange). 
    The positive class for frame-level is selected by taking the logical sum of the selected frames across the ground-truth video-level classes (in green hatched texture).}
    \label{fig:example_ASL}
\end{figure}

\subsubsection{Actionness Classifier}
To reduce the impact of irrelevant frames in a video, we propose the actionness classifier to identify which frames contain actions and which do not, denoted by \(p_a{(t)}\) for \(t \in \{1, \ldots, T\}\). 
However, since only video-level labels are available, it is challenging to learn this through strong label supervised learning. To address this issue, we select positive and negative frames based on the top-\(k\) predictions. 
As shown in Eq.~\ref{eq:top_k_selection}, we obtain the set of top-\(k\) predictions \(\mathcal{T}^c\). 
We use the video-level ground-truth classes to select positive frame $\mathcal{T}_\text{Pos}$ and negative frame $\mathcal{T}_\text{Neg}$ by taking the logical sum of the top-\(k\) predictions \(\mathcal{T}^c\) for the positive class across the ground-truth classes.
An example of the frame selection strategy is visualized in Figure~\ref{fig:example_ASL}.

Since the Actionness classifier is trained based on pseudo ground-truth derived from prediction probabilities, this leads to challenges during early training when action classification accuracy is poor and the top-\(k\) selections are not accurate. 
Consequently, the pseudo ground-truth can be very noisy, making it difficult for the model to converge. 
To address this problem, following~\cite{ma2021weakly} and~\cite{zhang2018generalized}, we employ the Generalized Cross Entropy Loss as:
\small
\begin{align}
\nonumber
\mathcal{L}_{\text{Actionness}} &= \frac{1}{|\mathcal{T}_{\text{Pos}}|} \sum_{t \in \mathcal{T}_{\text{Pos}}} \frac{1 - \left(p_a{(t)}\right)^q}{q} \\ &+ \frac{1}{|\mathcal{T}_{\text{Neg}}|} \sum_{t \in \mathcal{T}_{\text{Neg}}} \frac{1 - \left(1 - p_a{(t)}\right)^q}{q},
\label{eq:actionness}
\end{align}
\normalsize
where \(\mathcal{T}_{\text{Pos}}\) and \(\mathcal{T}_{\text{Neg}}\) denote the sets of positive and negative frames, respectively, and \(0 < q \leq 1\) determines the noise tolerance.

\subsection{Learning Objectives}
\label{subsec:learning_objectives}

The learning objectives of the proposed MultiASL model encompass three main components: View-level loss, Frame-level loss, and Actionness loss. 
The overall loss function is defined as:
\begin{equation}
\mathcal{L_{\text{Overall}}} = \mathcal{L}_{\text{View-level}} + \beta_1 \mathcal{L}_{\text{Frame-level}} + \beta_2 \mathcal{L}_{\text{Actionness}},
\label{eq:overall}
\end{equation}
where \(\beta_1\) and \(\beta_2\) are weighting factors. 
While the View-level loss facilitates multi-class prediction by aggregating all features within the multi-view videos, the fused features often contain irrelevant frames, which can hinder the learning process. 
The ASL aims to remove these irrelevant frames by training the Actionness classifier based on pseudo ground-truth obtained from frame-level predictions. This approach ensures that the model focuses on the most informative frames, thereby enhancing the robustness of the proposed method.

\begin{table*}[t]
\centering
\caption{Comparison of the proposed MultiASL and other methods. 
Best and second-best results are highlighted in \textbf{bold} and \underline{underlined} text, respectively.} 
{
  \begin{tabular}{lccccccccc}
    \toprule
    \multirowcell{3}[-4pt][l]{Method} & \multirowcell{3}[-4pt]{Backbone} & \multicolumn{8}{c}{View-level Fusion Strategy} \\
    \cmidrule(lr){3-10}
     & & \multicolumn{2}{c}{Max} & \multicolumn{2}{c}{Mean} & \multicolumn{2}{c}{Sum} & \multicolumn{2}{c}{Concat} \\ \cmidrule(lr){3-4} \cmidrule(lr){5-6} \cmidrule(lr){7-8} \cmidrule(lr){9-10}
     & & mAP${_C}$ & mAP${_S}$ & mAP${_C}$ & mAP${_S}$ & mAP${_C}$ & mAP${_S}$ & mAP${_C}$ & mAP${_S}$ \\
     
    \midrule
    MultiTrans~\cite{yasuda2022multi} & Resnet18 & 82.10 & 89.15 &  82.70 & 90.09 & 82.28 & 89.57 & 81.04 & 89.27 \\
    MultiTrans~\cite{yasuda2022multi} & Resnet34 & 82.93 & 89.47 & 83.57 & 90.88 & 83.03 & 90.74 & 81.41 & 89.44 \\
    MultiTrans~\cite{yasuda2022multi} & ViT & 84.47 & 90.64 & 84.73 & 91.08 & 82.56 & 86.66 & 80.40 & 87.71 \\
    Query2Label~\cite{liu2021query2label} & CLIP & 22.93 & 51.14 & 20.45 & 50.42 & 26.45 & 50.79 & 22.88 & 50.99 \\
    Query2Label~\cite{liu2021query2label} & ViT & 35.62 & 51.19 & 35.74 & 51.19 & 37.38 & 51.00 & 24.05 & 50.36 \\
    ConViT~\cite{d2021convit} & ViT & 81.92 & 89.20 & 80.27 & 87.54 & 81.96 & 88.93 & 80.09 & 88.52 \\
    TimeSformer~\cite{bertasius2021space} & ViT & 79.28 & 86.92 & 70.76 & 82.31 & 79.33 & 87.66 & 68.61 & 82.63  \\
    ViViT~\cite{arnab2021vivit} & ViT & 75.95 & 87.25 & 71.18 & 80.97 & 70.46 & 81.93 & 70.11 & 79.97  \\ 
    \midrule
    \multirowcell{3}[0pt][l]{MultiASL \\ (Proposed)} & Resnet18 & 86.12 & 90.61 & {\underline{86.71}} & {\underline{91.81}} & {\underline{86.22}} & 92.38 & {\underline{84.83}} & {\underline{90.11}} \\
    & Resnet34 & {\underline{86.55}} & \textbf{{93.06}} & 83.34 & 90.29 & 86.06 & \textbf{{93.32}} & 82.19 & 88.99  \\
    & ViT & \textbf{{88.05}} & {\underline{91.91}} & \textbf{{87.01}} & \textbf{{92.20}} & \textbf{{87.36}} & {\underline{92.95}} & \textbf{{86.33}} & \textbf{{90.16}} \\
    \bottomrule
  \end{tabular}
  \label{tab:main_results}
}
\end{table*}

\section{Experiments}
\label{sec:performance_evaluation}
% \begin{figure}[t]
%     \centering
%     \includegraphics[width=0.47\textwidth]{Fig/4_experiments/multi-view-example.pdf}
%     \caption{Visualization of synchronized data from the Multi-view MM-Office dataset~\cite{yasuda2022multi}.}
%     \label{fig:example_multiview}
% \end{figure}

\subsection{Experimental Conditions}
\subsubsection{Dataset}

% \begin{table}[t]
%     \centering
%     \caption{Description of the MM-Office dataset~\cite{yasuda2022multi} along with the number of events in the training set (\#Train) and the test set (\#Test).} 
%     \scalebox{0.90}{
%     \begin{tabular}{l|l|c|c}
%     \toprule
%     Class & Description & \#Train & \#Test  \\ \midrule
%     eat & Eating lunch in the desk. & \hspace{2pt} 44 & 20 \\ 
%     phone & Ringing a phone or cell phone. & \hspace{2pt} 32 & 20 \\
%     chat & Talking in-room. & 202 & 86 \\
%     tele & Remote communication via telephone. & \hspace{2pt} 90 & 38 \\
%     prepare & Setting up for a teleconference. & \hspace{2pt} 42 & 22 \\
%     meeting & Gathering at a large desk for a meeting. & \hspace{2pt} 44 & 20 \\
%     takeout & Take out something from shelf or box. & \hspace{2pt} 66 & 22 \\
%     handout & Handing out food or packages. & \hspace{2pt} 54 & 20 \\
%     enter & Entering and greeting. & \hspace{2pt} 90 & 38 \\
%     exit & Exiting the room. & \hspace{2pt} 44 & 20 \\
%     stand up & Stand up from a chair. & 103 & 41 \\
%     sit down & Sitting down on a chair. & 164 & 70 \\
%     \bottomrule
%     \end{tabular}
%     \label{tab:dataset}
%     }
% \end{table}

We utilize the MM-Office dataset~\cite{yasuda2022multi}, which comprises recordings from multiple cameras positioned at the 4 corners of an office room. 
Within this setting, 1 to 3 people engage in various daily work activities, classified into 12 distinct classes.
% categories as outlined in Table~\ref{tab:dataset}. 
% 
We use 2,816 videos captured from the 4 distributed cameras. 
Each video is annotated with multi-label tags, and the average length of the videos is 43.25 seconds. 
The dataset contains 1,392 events and has an unbalanced number of events across classes. 
To create the training and test sets, we use the Iterative Stratification strategy~\cite{sechidis2011stratification} for multi-label data with a 70:30 ratio, ensuring that all action classes are adequately represented in both subsets. 
% (see details in Supplementary materials).

For the experiments, we extract a fixed number of $T$ frames at a sampling rate of 2.5 FPS. 
During the training phase, from the total frames of the original video, we generate a perturbed sequence of frame indices for a given length $T$.
This sequence is based on uniform sampling, with random adjustments to ensure that the index covers the entire length, serving as a data augmentation method for robust training. 
During testing, we use uniform sampling to ensure that the index remains consistent across every test.

\subsubsection{Evaluation Metrics}
Following Kobayashi et al.~\cite{kobayashi2023two}, we evaluate performance using the two metrics below:
\begin{itemize}
    \item \(\text{mAP}_{C}\) (macro-averaged metric): Mean average precision is computed for each class and then aggregated across \textit{classes}. This is a primary metric in multi-label classification.    
    \item \(\text{mAP}_{S}\) (micro-averaged metric): Mean average precision is measured over \textit{samples}. This is a standard metric for single-label classification.
\end{itemize}

\subsubsection{Comparison Methods}
% To evaluate the effectiveness of the proposed method, we compare it with the following methods:
% \begin{itemize}
%     \item \textbf{MultiTrans~\cite{yasuda2022multi}:} Multi-view and multi-modal event detection using a Transformer-based multi-sensor fusion.
%     \item \textbf{Query2Label~\cite{liu2021query2label}:} Leverages Transformer decoders to query class labels for multi-label classification, utilizing CLIP~\cite{radford2021learning} features for vision and language queries.
%     \item \textbf{ConViT~\cite{d2021convit}:} Combines convolutional architecture and ViT strengths using gated positional self-attention. 
%     We modified ConViT with a Temporal Transformer Encoder to learn the temporal relationships in videos.
%     \item \textbf{TimeSformer~\cite{bertasius2021space} and ViViT~\cite{arnab2021vivit}:} Transformer-based approach for video classification using self-attention over spatial and temporal dimensions.
% \end{itemize}

To evaluate the effectiveness of the proposed method, we compare it with the following methods:
\begin{itemize}
    \item \textbf{MultiTrans~\cite{yasuda2022multi}:} Multi-view and multi-modal event detection by utilizing Transformer-based multi-sensor fusion. 
    It integrates data from distributed sensors through stacked Transformer blocks, effectively combining features from different viewpoints and modalities.
    \item \textbf{Query2Label~\cite{liu2021query2label}:} 
     Leverages Transformer decoders to query class labels for multi-label classification, utilizing Contrastive Language-Image Pre-Training (CLIP)~\cite{radford2021learning} features for vision and language queries.
    \item \textbf{ConViT~\cite{d2021convit}:} Improving ViT with soft convolutional inductive biases. 
    We modified ConViT with a Temporal Transformer Encoder to learn the temporal relationships in videos.
    \item \textbf{TimeSformer~\cite{bertasius2021space} and ViViT~\cite{arnab2021vivit}:}  Transformer-based architectures for video classification, applying self-attention mechanisms across spatial and temporal dimensions to capture complex video dynamics.
\end{itemize}

\subsubsection{Models \& Hyperparameters.}
We implement the proposal MultiASL method as detailed in Section~\ref{sec:methodology}\footnote{The source code is available at \url{https://github.com/thanhhff/MultiASL/}.}. 
The input to the model consists of 4 views captured from cameras with a fixed frame of \(T = 50\) and a resolution of \(3 \times 224 \times  224 \). 
For the Spatial Encoder, we employ different backbones including ResNet18~\cite{he2016deep}, ResNet34~\cite{he2016deep}, and ViT~\cite{dosovitskiy2020image}. 
For the Temporal Transformer Encoder, we use 1 Transformer Encoder~\cite{vaswani2017attention} layer with 4 heads, each having a dimension of 128. 
The balancing parameter \(\alpha\) and temperature parameter \(\gamma\) in Eq.~\ref{eq:loss_view_level} are set to 1 and 4, respectively.
The noise tolerance \(q\) in Eq.~\ref{eq:actionness} is set to 0.7, where \(q\) being close to 1 makes the model more tolerant to deviations from the ground-truth. 
The parameters \(\beta_1\) and \(\beta_2\) in Eq.~\ref{eq:overall} are both set to 1 for simplicity. 
The parameter \(k\) for selecting the top-\(k\) probabilities is set to \(k = T/8\).
We employ multiple methods for View-level fusion, including max pooling, mean pooling, sum, and concatenation, while for Frame-level fusion, we use max pooling. 
For optimization, we use the Adam optimizer~\cite{kingma2014adam} with an initial learning rate set at \(10^{-4}\) and a weight decay of \(5.0\times 10^{-4}\)  until convergence. The maximum number of epochs is fixed at 50 with a batch size of 8. 
All experiments are performed on a machine equipped with an AMD EPYC 7402P 24-core processor and an NVIDIA A6000 GPU.

%%%%%%%%%%%%%

\begin{table*}[t]
\centering
\caption{Multi-label single-view action recognition. Average precision of each class is reported, with the best results are highlighted in \textbf{bold}.} 
\vspace{-5pt}
{
  \resizebox{1.0\linewidth}{!}{\begin{tabular}{cccccccccccccc}
    \toprule
     \multirowcell{1}[-0pt][c]{Setting} & Eat & Tele & Chat & Meeting & Takeout & Prepare & Handout & Enter & Exit & Stand up & Sit down & Phone & Average \\
    \midrule
    All view & \textbf{90.67} & 98.22 & \textbf{96.62} & \textbf{97.84} & 98.27 & \textbf{98.47} & \textbf{56.66} & 98.80 & \textbf{95.37} & \textbf{72.52} & \textbf{77.66} & 75.53 & \textbf{88.05} \\
    View 1 & 79.77 & \textbf{98.96} & 96.06 & 97.08 & 97.54 & 98.02 & 45.34 & \textbf{99.17} & 93.95 & 52.21 & 73.13 & 65.60 & 83.07 \\
    View 2 & 83.96 & 98.07 & 93.98 & 92.25 & 98.09 & 96.27 & 46.21 & 98.53 & 87.26 & 58.26& 71.62& \textbf{79.21} & 83.64 \\
    View 3 & 71.90 & 93.12 & 82.68 & 54.63 & \textbf{99.25} & 86.71 & 39.22 & 95.32 & 85.58 & 49.07 & 62.12 & 70.21 & 74.15 \\
    View 4 & 47.06 & 85.23 & 68.81 & 94.65 & 53.48 & 79.93 & 25.28 & 34.42 & 29.89 & 55.55 & 56.68 & 73.90 & 58.74 \\
    \bottomrule
  \end{tabular}
  \label{tab:ablation_single}
}}
\end{table*}

\begin{table}[t]
\centering
\caption{Impact of different loss components. Best results are highlighted in \textbf{bold}.} 
\vspace{-5pt}
{
  \begin{tabular}{ccccc}
    \toprule
     \multirowcell{2}[-0pt][c]{View-level\\ loss} & \multirowcell{2}[-0pt][c]{Frame-level\\ loss} & \multirowcell{2}[-0pt][c]{Actioness \\ loss} & \multirowcell{2}[-0pt][c]{mAP${_C}$} & \multirowcell{2}[-0pt][c]{mAP${_S}$} \\
    % \cmidrule{3-5}
     & & &   \\
    \midrule
    $\checkmark$ & $\checkmark$ & $\checkmark$ & \textbf{88.05} & \textbf{91.91}  \\
    $\checkmark$ & $\checkmark$ & & 85.90 & 90.99  \\
    $\checkmark$ &  &  & 84.04 & 89.90 \\
     & $\checkmark$ & $\checkmark$ & 84.30 & 89.12  \\
     & $\checkmark$ &  & 81.65 & 87.82  \\
    \bottomrule
  \end{tabular}
  \label{tab:ablation_loss}
}
\end{table}

%%%%%%%%%%%%%

\subsection{Results}

Table~\ref{tab:main_results} shows the results of the proposed MultiASL compared to other methods. 
 It demonstrates superior performance across various backbone architectures (i.e., Resnet18, Resnet34, and ViT) and view-level fusion strategies (i.e., max pooling, mean pooling, sum, and concatenation) in terms of mAP$_{C}$ and mAP$_{S}$. 
Notably, MultiASL with ResNet34 and ViT backbones consistently achieves the highest mAP. 
View-level fusion with max-pooling achieves the best results compared to other fusion strategies. 
With the ViT backbone, MultiASL achieves the best results with mAP$_{C}$ of 88.05\% and mAP$_{S}$ of 91.91\% using max pooling fusion, outperforming the MultiTrans, Query2Label, ConViT, TimeSformer, and ViViT models. 
These results demonstrate the robustness and effectiveness of MultiASL, with the ASL effectively selecting the action frames while reducing irrelevant information, thus enhancing accuracy.

\subsection{Ablation Study}
We conduct a series of excision experiments as part of an ablation study to evaluate the effectiveness of the MultiASL, using ViT as the spatial feature extractor and max-pooling for view-level fusion.

\subsubsection{Multi-label Single-view Action Recognition.}
To demonstrate the effectiveness of the proposed multi-view fusion, we evaluated single-view input settings as shown in Table~\ref{tab:ablation_single}. 
The results indicate that single-view settings generally performed worse compared to multi-view fusion. 
Views 1 and 2 provided relatively higher accuracy due to their advantageous positioning and angles, capturing the most frequently occurring actions. 
Conversely, Views 3 and 4 exhibited lower accuracy due to the distributed camera setup, where views overlap but do not focus around the target, causing some actions not visible from those views. 
The results also show that instantaneous actions such as ``Stand up'', ``Sit down'', and ``Handout'' were challenging to detect, suggesting these actions might occur in brief moments easily missed from a single view. 
Moreover, View 2 showed superior results for the ``Phone'' action due to its proximity to the desk phone. 
On the other hand, View 4 performed worst in the ``Enter'' and ``Exit'' actions because the camera's direction was not directly towards the door.
These results demonstrate that integrating all views enhanced accuracy by allowing information from different views to complement each other.

\subsubsection{Impact of Different Loss Components.}
Table~\ref{tab:ablation_loss} shows the impact of different loss components on the performance of the proposed method. 
The full model achieved the highest performance. 
Removing the Actionness loss resulted in a slight performance drop, indicating its importance in improving prediction accuracy by identifying which frames contain actions. 
Further removing the Frame-level loss caused a greater drop in performance, demonstrating the critical role of ASL in the overall accuracy. 
When only the Frame-level and Actionness losses were used, the model still performed well, but it is clear that the View-level loss is essential for achieving optimal performance. 
Lastly, the absence of both the View-level and Actionness losses resulted in the lowest performance, due to the difficulty in selecting top-$k$ predictions to determine the video-level class.

\section{Conclusion and Discussion}
\label{sec:conclusion}
This study proposed the MultiASL method for enhancing multi-label multi-view action recognition using only video-level labels. 
By employing a Multi-view Spatial-Temporal Transformer video encoder and Action Selection Learning, the proposed approach effectively identifies and integrates relevant frames from multiple viewpoints. 
Various fusion strategies were explored to optimize the integration of information from different views. 
The experimental results on the real-world office environments offered by the MM-Office dataset demonstrated the robustness and accuracy of the MultiASL method, significantly outperforming existing methods.

\noindent \textbf{Limitations and future work:}
In this study, we experimented with fixed hyper-parameters (i.e., top-$k$ probabilities), leading to the need for future research to explore the impact of varying these hyper-parameters to optimize performance. 
Additionally, MultiASL relies solely on camera visual information, potentially missing contextual information from other modalities such as audio. 
Integrating multi-modal data could significantly enhance the robustness and accuracy of multi-label multi-view action recognition.
% Moreover, investigating advanced techniques for handling weak labels and improving model training efficiency are also crucial directions for future research.

\begin{acks}
This work was partly supported by JSPS KAKENHI JP21H03519 and JP24H00733. 
\end{acks}

\bibliographystyle{ACM-Reference-Format}
\bibliography{references}
\end{document}